\documentclass{opt2021} 
\usepackage{times}

\usepackage{amsmath,amsfonts,bm}

\def\eqref#1{equation~\ref{#1}}

\def\1{\bm{1}}

\DeclareMathAlphabet{\mathsfit}{\encodingdefault}{\sfdefault}{m}{sl}
\SetMathAlphabet{\mathsfit}{bold}{\encodingdefault}{\sfdefault}{bx}{n}

\usepackage{wrapfig}
\usepackage{hyperref}
\usepackage{url}

\usepackage[utf8]{inputenc}
\usepackage{microtype}
\usepackage{booktabs} 

\usepackage{hyperref}

\usepackage{amsmath}
\usepackage{amssymb}
\usepackage{algorithm}               
\usepackage{algorithmic}             
\usepackage{caption}
\usepackage{multirow}                
\newcommand{\tabincell}[2]{\begin{tabular}{@{}#1@{}}#2\end{tabular}}
\title[Community-based Layerwise Distributed Training of Graph Convolutional Networks]{Community-based Layerwise Distributed Training of Graph Convolutional Networks}

\optauthor{%
\Name{Hongyi Li} \Email{lihongyi@stu.xidian.edu.cn}\\
\addr The State Key Laboratory of Integrated Service Networks, Xidian University, Xi'an, Shaanxi, China, 710071
\AND
\Name{Junxiang Wang} \Email{junxiang.wang@emory.edu}\\
\addr Department of Computer Science and Informatics,
Emory University, Atlanta, Georgia, USA, 30030
\AND
\Name{Yongchao Wang} \Email{ychwang@mail.xidian.edu.cn}\\
\addr The State Key Laboratory of Integrated Service Networks, Xidian University, Xi'an, Shaanxi, China, 710071
\AND
\Name{Yue Cheng} \Email{yuecheng@gmu.edu}\\
\addr Department of Computer Science, George Mason University, Fairfax, Virginia, USA, 22030
\AND
\Name{Liang Zhao} \Email{liang.zhao@emory.edu}\\
\addr  Department of Computer Science and Informatics,
Emory University, Atlanta, Georgia, USA, 30030
}

\newtheorem{problem}{Problem}
\begin{document}

\maketitle

\begin{abstract}
\indent The Graph Convolutional Network (GCN) has been successfully applied to many graph-based applications.
Training a large-scale GCN model, however, is still challenging: Due to the node dependency and layer dependency of the GCN architecture, a huge amount of computational time and memory is required in the training process. In this paper, we propose a parallel and distributed GCN training algorithm based on the Alternating Direction Method of Multipliers (ADMM) to tackle the two challenges simultaneously. We first split GCN layers into independent blocks to achieve layer parallelism. Furthermore, we reduce node dependency by dividing the graph into several dense communities such that each of them can be trained with an agent in parallel.  Finally, we provide solutions for all subproblems in the community-based ADMM algorithm. Preliminary results demonstrate that our proposed community-based ADMM training algorithm can lead to more than triple speedup while achieving the best performance compared with state-of-the-art methods.

\end{abstract}
\section{Introduction}
Graphs are prevalent structures in various real-world applications including social networks \citep{qiu2018deepinf}, recommender systems \citep{ying2018graph}, and biology and chemistry networks \citep{duvenaud2015convolutional}, which has attracted much attention from the deep learning community. Graph Convolutional Network (GCN) is one of the leading graph neural network architectures due to its impressive performance on many downstream tasks (e.g. node classification, link prediction, and graph classification) \citep{kipf2016semi}. However, it is challenging to train  GCN efficiently due to two difficulties: \textbf{1) Node dependency.} The GCN needs to propagate information among nodes through node interactions in the graph. This means that the loss for each node depends on a large number of neighboring nodes. Such dependency becomes more complex as the GCN goes deeper. \textbf{2). 
Layer dependency.} The interactions between nodes are transmitted through layers. Therefore for the backpropagation algorithm, the gradient of node interactions in one layer relies on that in previous layers. Because of node dependency and layer dependency, training a large-scale GCN requires a lot of computational time and memory: node representations in different layers are required to be updated in sequential, and all of them are required to be stored in the CPU memory. \\
\indent In order to address these two challenges simultaneously, in this paper, we propose a distributed and parallel GCN training algorithm based on the Alternating Direction Method of Multipliers (ADMM). This is because ADMM has attained great achievements in training deep neural networks in parallel via layer splitting \cite{Wang2020toward}. Specifically, it breaks a series of layers into independent blocks, in order to alleviate layer dependency. Moreover, the complexity of node dependency can be reduced significantly (i.e. from multi-layer level to one-layer level). Apart from layer splitting, we also partition a graph into independent communities: unlike previous works such as Cluster-GCN \citep{chiang2019cluster}, which remove inter-community connections and thus degrade performance, we maintain node connections, which contain the first-order and second-order neighboring information, and realize parallel training by multiple agents without performance loss. Preliminary experiments on two benchmark datasets demonstrate that our proposed community-based ADMM algorithm leads to more than triple speedup and achieves superior performance compared with state-of-the-art optimizers such as SGD and Adam.

\section{Problem Formulation}

We formulate the GCN training problem in this section. Let $\mathcal{G}(\mathcal{V}; \mathcal{E})$ be an undirected and unweighted graph, where $\mathcal{V}$  and $ \mathcal{E}$ are sets of nodes and edges, respectively. $N=\vert \mathcal{V}\vert$ is the number of nodes. $A, {D}\in \mathbb{R}^{n \times n}$ are an adjacency matrix and a degree matrix, respectively. Then the GCN training problem is formulated mathematically as follows:

\begin{problem}
\vspace{-0.2cm}
\small
\label{prob: problem 1}
\begin{align*}
   &\min\nolimits_{\textbf{W},\textbf{Z}} \quad R(Z_L,Y) \quad s.t. \quad {Z}_l = f_l(\tilde{A}Z_{l-1}W_{l}) \quad (l=1, \cdots, L-1), \quad {Z}_L = \tilde{A}Z_{L-1}W_{L},
\end{align*}
\normalsize
\end{problem}
 \begin{wraptable}{r}{0.5\linewidth}
\scriptsize
\centering
 \begin{tabular}{cc}
 \hline
 Notations&Descriptions\\ \hline
 $L$& Number of layers.\\
 $N$& Number of nodes.\\
 $A$ & The adjacency matrix of a graph.\\
 $D$ & The degree matrix of a graph.\\
 $W_l$& The weight matrix for the $l$-th layer.\\
 $f_l(\cdot)$& The nonlinear activation function for the $l$-th layer.\\
 $Z_l$& The output for the $l$-th layer.\\
 $Z_0$& The input feature matrix for the neural network.\\
 $Y$& The predefined label matrix.\\
 $R(Z_L,Y)$& The risk function for the $L$-th layer.\\
 $C_l$& The number of neurons for the $l$-th layer.\\
\hline
  \end{tabular}
  \vspace{-0.3cm}
  \captionof{table}{Important Notations}
   \label{tab:notation}
 \end{wraptable}
\normalsize
 
where $\mathbf{W}=\{{W_l}\}_{l=1}^{L}, \mathbf{Z}=\{{Z_l}\}_{l=1}^{L}$, and ${\tilde{A}} = ({D}+{I})^{-1/2}({A}+{I})({D}+{I})^{-1/2}$ is a normalized adjacency matrix. $Z_0 \in \mathbb{R}^{n \times C_0}$ is an input feature matrix, where each row corresponds to an input feature vector of a node, and $C_l$ is the number of hidden units for the $l$-th layer. $W_l \in \mathbb{R}^{C_{l-1} \times C_{l}}$ and  $Z_l \in \mathbb{R}^{n \times C_{l}}$ are the weight matrix and the output for the $l$-th layer, respectively. ${Y}\in\mathbb{R}^{n\times C_L}$ is the pre-defined label matrix, and $C_L$ is the number of node classes.   $f_l$ is a non-linear activation function for the $l$-th layer (e.g., ReLU). $R(\cdot)$ is a risk function such as the cross-entropy loss. Problem \ref{prob: problem 1} is difficult to solve due to nonlinear constraints $Z_l=f_l(\tilde{A}Z_{l-1}W_l)$.  Therefore we relax it to  Problem \ref{prob: problem 2} as follows:
\begin{problem}
\vspace{-0.5cm}
\small
\label{prob: problem 2}
\begin{equation*}
    \begin{aligned}
   &\min\nolimits_{\mathbf{W},\mathbf{Z}} \quad R({Z}_L,{Y}) +  \frac{\nu}{2}\sum_{l=1}^{L-1}\Vert{Z}_l - f_l(\tilde{{A}}{Z}_{l-1}{W}_{l})\Vert_F^2 \quad
  s.t. \quad {Z}_L = \tilde{{A}}{Z}_{L-1}{W}_{L},
   \end{aligned}
\end{equation*}
\vspace{-0.5cm}
\normalsize
\end{problem}
\begin{wrapfigure}{r}{\linewidth}
    \includegraphics[width=\linewidth]{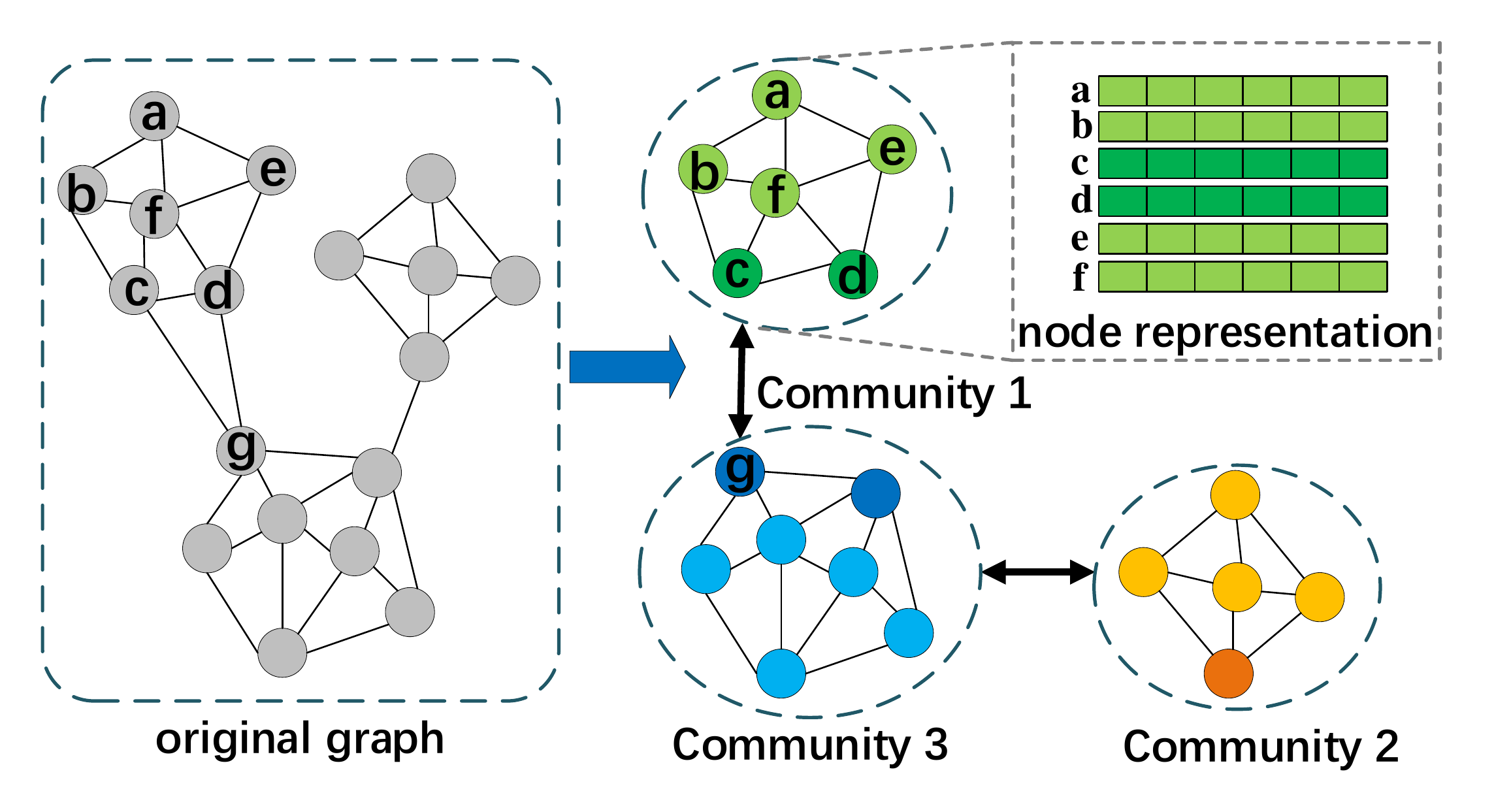}
    \centering
    \caption{Illustrations of the graph partition: a graph is split into three community. }
    \label{fig:gnn_community}
\end{wrapfigure}
where $\nu >0$ is a tuning parameter. Note that when $\nu \longrightarrow \infty$,  Problem \ref{prob: problem 2} approaches Problem \ref{prob: problem 1}.\\
\indent Many graph problems such as node classification and link prediction are 
applied in large-scale scenarios (e.g. social networks), where the adjacency matrix cannot fit in memory. Motivated by 
Cluster-GCN \cite{chiang2019cluster}, we divide the graph $\mathcal{G}$ into $M$ communities by METIS\cite{Karypis1998A}, where $\mathcal{V}=\bigcup\nolimits_{m=1}^M \mathcal{V}_m$, $\mathcal{V}_m\cap \mathcal{V}_j=\emptyset (1 \leq m<j\leq M)$. $n_m=\vert \mathcal{V}_m\vert$ is the number of nodes in the $m$-th community. Each community can be fed to an independent agent for distributed training.
A community index set neighboring the $m$-th community is defined as $\mathcal{N}_m=\{ i|\exists (u,v)\in \mathcal{E},u\in \mathcal{V}_m, v\in \mathcal{V}_i, i \neq m\}$. Figure \ref{fig:gnn_community} illustrates the partition of communities, where a graph is split into three communities.
$\mathcal{N}_1= \{3\}$ because $c, d$ in community 1 are commected to $g$ in community 3.\\

\indent We split  $\tilde{{A}}$ according to the partition of $\mathcal{G}$ as follows:
\small{\[\tilde{{A}} = \left[ {\begin{array}{*{20}{c}}
{{\tilde{{A}}_{1,1}}}& \cdots &{{\tilde{{A}}_{1,M}}}\\
 \vdots & \ddots & \vdots \\
{{\tilde{{A}}_{M,1}}}& \cdots &{{\tilde{{A}}_{M,M}}}
\end{array}} \right],\]}
\normalsize
where $\tilde{{A}}_{m,m} \in \mathbb{R}^{n_m\times n_m}$ represents the adjacency matrix of the $m$-th community, and $\tilde{A}_{m,j}$ defines the topology between the $m$-th and $j$-th communities.
Accordingly, $Z$ and $Y$ are partitioned as
${Z}_l = [{Z}_{l,1}^T,\cdots,{Z}_{l,M}^T]^T(l=1, \cdots, L)$ and
$Y = [Y_1^T, \cdots, Y_M^T]^T $.
Then Problem \ref{prob: problem 2} is equivalently transformed to the following:

\begin{problem}
\label{prob: problem 3}
\small
    \begin{align*}
   &\min\nolimits_{\mathbf{W},\mathbf{Z}}  \sum_{m=1}^M R(Z_{L,m}, Y_{m})
   +\!\frac{\nu}{2}\!\sum_{l=1}^{L-1}\!\sum_{m=1}^M\!\|{Z}_{l,m}\! -\! f_l((\tilde{{A}}_{m,m}{Z}_{l-1,m}\!+\!\sum_{r \in \mathcal{N}_m}\tilde{{A}}_{m,r}{Z}_{l-1,r}){W}_{l})\|_F^2 \\
   &s.t. \quad {Z}_{L,m}=(\tilde{{A}}_{m,m}{Z}_{L-1,m}+\sum_{r \in \mathcal{N}_m}\tilde{{A}}_{m,r}{Z}_{L-1,r}){W}_{L} \ (m=1,\cdots,M).
   \end{align*}
\normalsize
\vspace{-0.2cm}
\end{problem}

\vspace{-0.5cm}
\begin{algorithm}[H]
\small
\caption{The Community-based  ADMM Algorithm\label{algo:community_ADMM_GCN}}
\begin{algorithmic}[1]
 \REQUIRE ${Y}$, ${Z}_0$, $\nu, \rho$.
\ENSURE ${Z}_{l,m}$, ${W}_{l}, l=1,\cdots, L,  m=1,\cdots, M$.
\STATE \textbf{Initialize:} $k=0$.
\WHILE{ ${W}_{l}^k, {Z}_{l,m}^k$ not converged} 
\STATE Update ${W}_{l}^{k+1}$  for different $l$ in parallel.
\STATE Update ${Z}_{l,m}^{k+1}$ for different $l$ and $m$ in parallel.
\STATE Update ${U}_m^{k+1}$ for different $m$ in parallel.
\ENDWHILE
\end{algorithmic}
\end{algorithm}
\normalsize
\vspace{-0.5cm}
\section{The community-based ADMM Algorithm}
\vspace{-0.1cm}
\indent In this section, we propose the ADMM algorithm to solve Problem \ref{prob: problem 3}.
The augmented Lagrangian is formulated mathematically as follows:
\small
\vspace{-0.2cm}
\begin{align}
   \mathcal{L}_{\rho}(\mathbf{W},\mathbf{Z},\mathbf{U})&= \sum_{m=1}^M R(Z_{L,m}, Y_{m})\!
   +\!\frac{\nu}{2}\!\sum_{l=1}^{L-1}\!\sum_{m=1}^M\!\|{Z}_{l,m}\! -\! f_l((\tilde{{A}}_{m,m}{Z}_{l-1,m}\!+\!\sum_{r \in \mathcal{N}_m}\tilde{{A}}_{m,r}{Z}_{l-1,r}){W}_{l})\|_F^2\!\nonumber\\
   &+\!\sum_{m=1}^M(\langle {U}_{m}, ({Z}_{L,m} \!-\! (\tilde{{A}}_{m,m}{Z}_{L-1,m}\!+\!\sum_{r \in \mathcal{N}_m }\tilde{{A}}_{m,r}{Z}_{L-1,r}){W}_{L})\rangle \nonumber
   \\&\!+\! \frac{\rho}{2}\|{Z}_{L,m}\!-\! (\tilde{{A}}_{m,m}{Z}_{L-1,m}\!+\!\sum_{r \in \mathcal{N}_m}\!\tilde{{A}}_{m,r}{Z}_{L-1,r}){W}_{L}\|_F^2), \label{eq:lagrangian_cluster}
\end{align}
\normalsize
where  $\rho>0$ is a penalty parameter, and $\mathbf{U}=\{U_m\}_{m=1}^M$ are Lagrangian multipliers. The ADMM algorithm to solve \eqref{eq:lagrangian_cluster} is shown in Algorithm \ref{algo:community_ADMM_GCN}.  Specifically, Lines 3 and 4 update $\mathbf{W}$ (i.e. layerwise training) and $\mathbf{Z}$ (i.e. community-wise training) in parallel, respectively, and Line 7 updates $\mathbf{U}$. All subproblems are discussed in detail as follows.
For the sake of simplicity, we define
 \small
\begin{align*}
     \phi({W}_{l},  {Z}_{l-1}, {Z}_{l})&
     \triangleq \frac{\nu}{2}\sum_{m=1}^M\|{Z}_{l,m}\!-\! f_l((\tilde{{A}}_{m,m}{Z}_{l-1,m}\!+\!\sum_{r \in \mathcal{N}_m}\!\tilde{{A}}_{m,r}{Z}_{l-1,r}){W}_{l})\|_F^2\\&=\frac{\nu}{2}\|{Z}_{l} - f_l(\tilde{{A}}{Z}_{l-1}{W}_{l})\|_F^2\ (l=1,\cdots,L-1),
\end{align*}
\vspace{-2pt}
\normalsize
and
\small
\begin{align*}
     &\phi({W}_{L},  {Z}_{L-1}, {Z}_{L},  {U}) \triangleq \sum_{m=1}^M(\langle U_m, {Z}_{L,m} \!-\!(\tilde{{A}}_{m,m}{Z}_{l-1,m}+\sum_{r \in \mathcal{N}_m}\tilde{{A}}_{m,r}{Z}_{l-1,r}){W}_{l}\rangle\\
     &\!+\!\frac{\rho}{2}\|{Z}_{L,m}\!-\! (\tilde{{A}}_{m,m}{Z}_{l-1,m}\!+\!\sum_{r \in \mathcal{N}_m}\tilde{{A}}_{m,r}{Z}_{l-1,r}){W}_{l}\|_F^2)=\langle{U}, {Z}_L\!-\! \tilde{{A}}{Z}_{L-1}{W}_{L}\rangle\!+\! \frac{\rho}{2}\|{Z}_L\!-\! \tilde{{A}}{Z}_{L-1}{W}_{L}\|_F^2,
\end{align*}
\vspace{-5pt}
\normalsize
where
$U = [U_1^T, \cdots, U_M^T]^T$.

\subsection{Update $W_l^{k+1}$}
The variable  $W_l^{k+1}$ is updated on agent $M+1$ as follows:
\small
\begin{align*}
    W^{k+1}_l&\leftarrow \arg\min\nolimits_{W_l} \mathcal{L}_\rho(\textbf{W},\textbf{Z}^{k},\textbf{U}^{k}) =\arg\min\nolimits_{W_l} \begin{cases} \phi({W}_{l},  {Z}_{l-1}^k, {Z}_{l}^k)
    &l<L\\\phi({W}_{L},  {Z}_{L-1}^k, {Z}_{L}^k,  {U}^k)&l=L
    \end{cases}
\end{align*}
\normalsize

 Agent $m(m<M+1)$ needs to send $Z_{l,m}^{k}(l>0)$ and $U_{m}^{k}$ to agent $M+1$ in advance to form  ${Z}_{L-1}^k, {Z}_{L}^k$, and  ${U}^k$.  Furthermore,  solving ${W}_{l}^{k+1}$ requires the inverse of $\tilde{A}$, which is usually not inversible.  To handle this, we apply the quadratic approximation \citep{Wang2019} as follows:
 \small
\begin{equation}\label{eq:update w}
{W}_{l}^{k+1}\leftarrow \arg\min\nolimits_{{W}_{l}} {P}_{l}({W}_{l};\tau_{l}^{k+1}),
\end{equation}
\normalsize
where 
\small
\begin{align*}
&{P}_l({W}_{l};\tau_l^{k+1})\!=\!\begin{cases}
\phi({W}_{l}^k, {Z}_{l-1}^{k}, {Z}_{l}^{k})
\!+\!\langle\nabla_{{W}_{l}^k}\phi({W}_{l}^k,{Z}_{l-1}^{k},  {Z}_{l}^{k}),{W}_{l}-{W}_{l}^k\rangle
\!+\!\frac{\tau_l^{k+1}}{2}\|{W}_{l}\!-\!{W}_{l}^k\|_F^2 ,l<L \\
\phi({W}_{L}^k, {Z}_{L-1}^{k}, {Z}_{L}^{k}, U^k)
\!+\!\langle\nabla_{{W}_{L}^k}\phi({W}_{L}^k, {Z}_{L-1}^{k}, {Z}_{L}^{k}, U^k),{W}_{L}\!-\!{W}_{L}^k\rangle
\!+\!\frac{\tau_l^{k+1}}{2}\|{W}_{L}\!-\!{W}_{L}^k\|_F^2,\\
\end{cases}
 \end{align*}
 \normalsize

 and $\tau_l^{k+1}>0$ is a parameter that should satisfy:
 \small
\begin{align*}
{P}_l({W}_{l}^{k+1};\tau_l^{k+1}) \geq
\begin{cases}
\phi({W}_{l}^{k+1}, {Z}_{l-1}^{k}, {Z}_{l}^{k}), l<L \\
\phi({W}_{L}^{k+1}, {Z}_{L-1}^{k}, {Z}_{L}^{k}, U^k),l=L.\\
\end{cases}
 \end{align*}
\normalsize
The solution to \eqref{eq:update w} is:
\small
\begin{align*}
{W}_{l}^{k+1}
\!\leftarrow\!
\begin{cases}
\!{W}_{l}^{k}\!-\!\nabla_{{W}_{l}^k}\phi({W}_{l}^k, {Z}_{l-1}^{k}, {Z}_{l}^{k})/\tau_l^{k+1}, l\!<\!L\\
\!{W}_{L}^{k}\!-\!\nabla_{{W}_{L}^k}\phi({W}_{L}^k, {Z}_{L-1}^{k}\!,\! {Z}_{L}^{k},U^k)/\tau_L^{k+1},l\!=\!L.
\end{cases}
\end{align*}
\normalsize
Obviously, $W_l^{k+1}$ for different layers can be updated in parallel.
\subsection{Update $Z_{l,m}^{k+1}$}
The update of  $Z_{l,m}^{k+1}$ resembles that of  $W_{l}^{k+1}$. Due to space limit,  details are given in Appendix \ref{sec:update z}.
\subsection{ Update $U_m^{k+1}$}
The variable   $U_{m}^{k+1}$ is updated as follows:
\small
\begin{equation}\label{eq:update u}
U_m^{k+1}\leftarrow
U_m^{k}+\rho({Z}_{L,m}^{k}-(\sum_{r \in \mathcal{N}_m\cup{\{m\}}}p_{L-1,r\rightarrow m}^{k})),
\end{equation}
\normalsize
where  $p_{L-1,r\rightarrow m}^{k}$ is defined in  Appendix \ref{sec:update z}.
\vspace{-5pt}

\section{Experiments}
\vspace{-0.1cm}
In this section, we evaluate the performance of the proposed community-based ADMM algorithm using two benchmark datasets. Four state-of-the-art optimizers are used as comparison methods in terms of both accuracy and speedup. All experiments were conducted on a 64-bit machine with Intel(R) Xeon(R) Silver 4110 CPU and 64GB RAM. The statistics of two benchmark datasets are shown in Table \ref{tab:dataset}. 
\subsection{Speedup}\label{sec:speedup}
In this experiment, we investigate the speedup of the proposed ADMM algorithm on a two-layer GCN model with 1000 hidden units. The activation function was set to the Rectified Linear Unit (ReLU). The loss function was the cross-entropy loss. The running time per epoch was an average of 50 epochs. $\rho$ and $\nu$ were both set to $10^{-3}$ for Amazon Computers and $10^{-4}$ for Amazon Photo. Specifically, in the Serial ADMM algorithm, we used only one community, and the two layers were trained sequentially; while in the Parallel ADMM algorithm, we divided the original graph into 3 communities that were trained by 3 agents simultaneously, plus applied a layer parallelism scheme. \\
\indent The training and communication time, as well as speedup, were listed in Table \ref{tab:time} on Amazon Computers and Amazon Photo. The training time on both datasets was reduced by more than $80\%$. Although the Parallel ADMM involves additional time for communication among agents, it is still around 2$\times$ faster than the Serial ADMM method, which demonstrates the effectiveness of the proposed community-based algorithm.

\subsection{Accuracy}
\begin{figure*}
    \begin{minipage}{0.24\linewidth}
    \includegraphics[width=\linewidth]{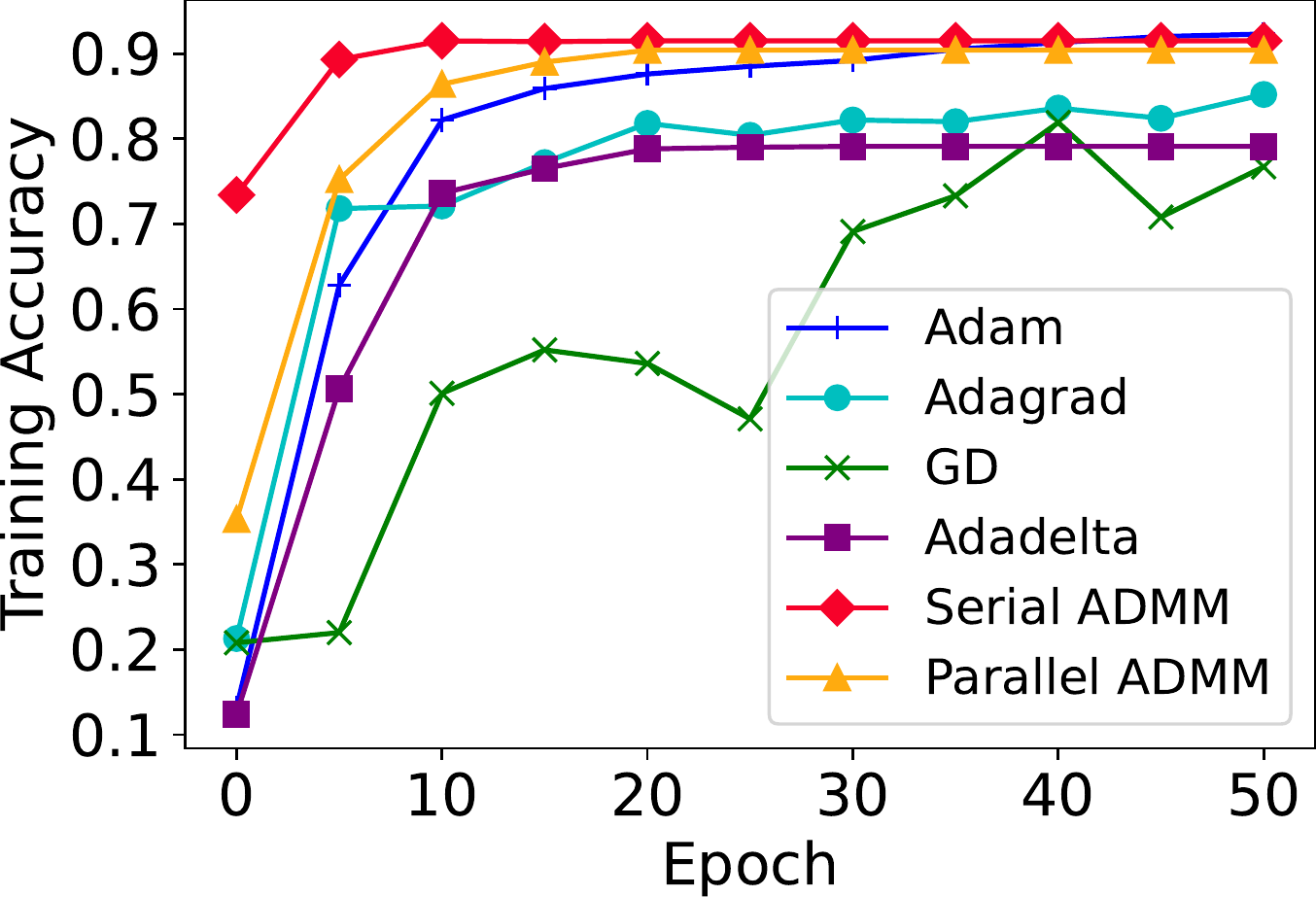}
    \centerline{ \tabincell{c}{(a). Training accuracy\\ for Amazon Computers.}}
    \end{minipage}
    \hfill
    \begin{minipage}{0.24\linewidth}
    \includegraphics[width=\columnwidth]{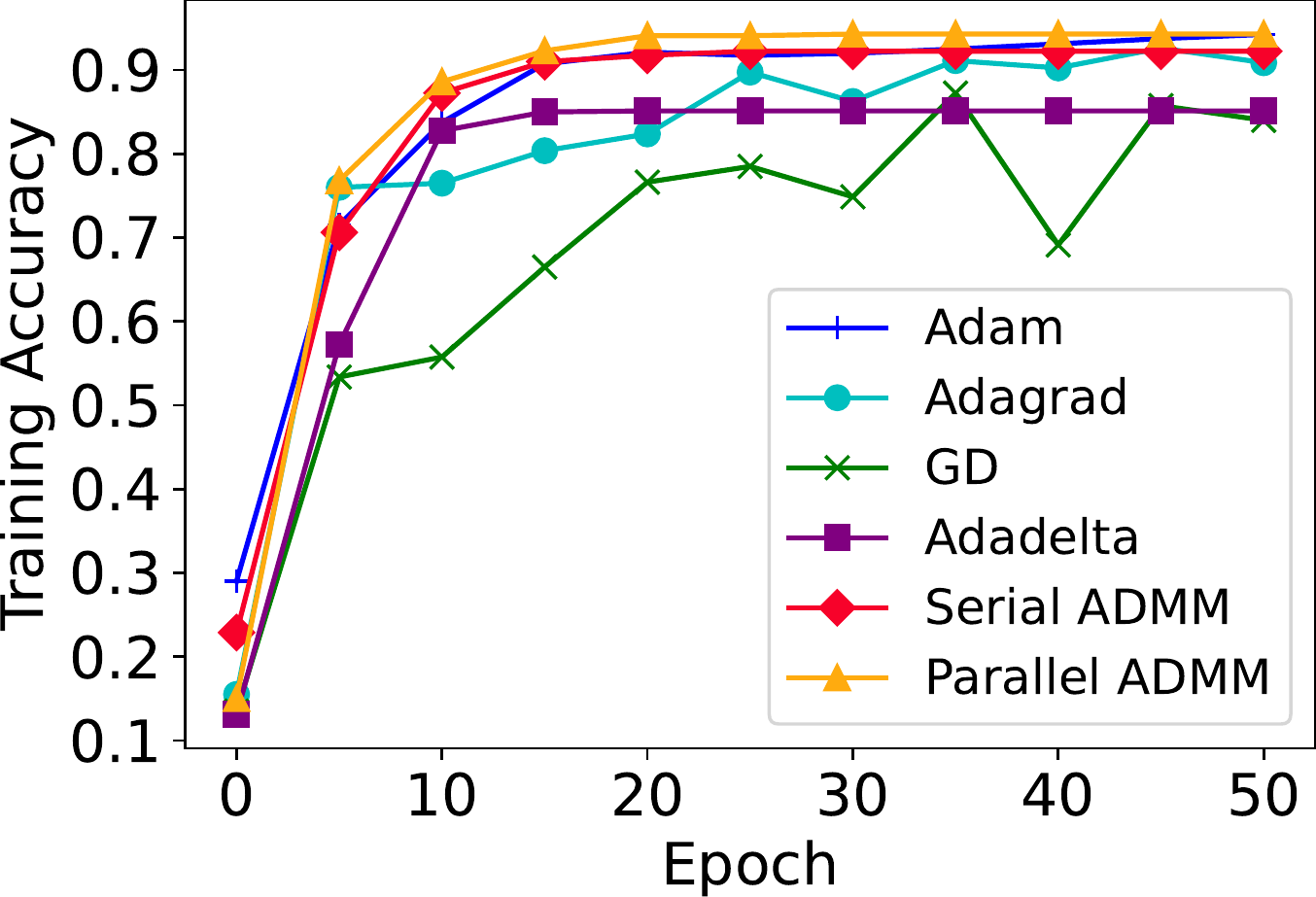}
    \centerline{\tabincell{c}{(b). Training accuracy\\ for Amazon Photo.}}
    \end{minipage}
    \hfill
    \begin{minipage}{0.24\linewidth}
    \includegraphics[width=\linewidth]{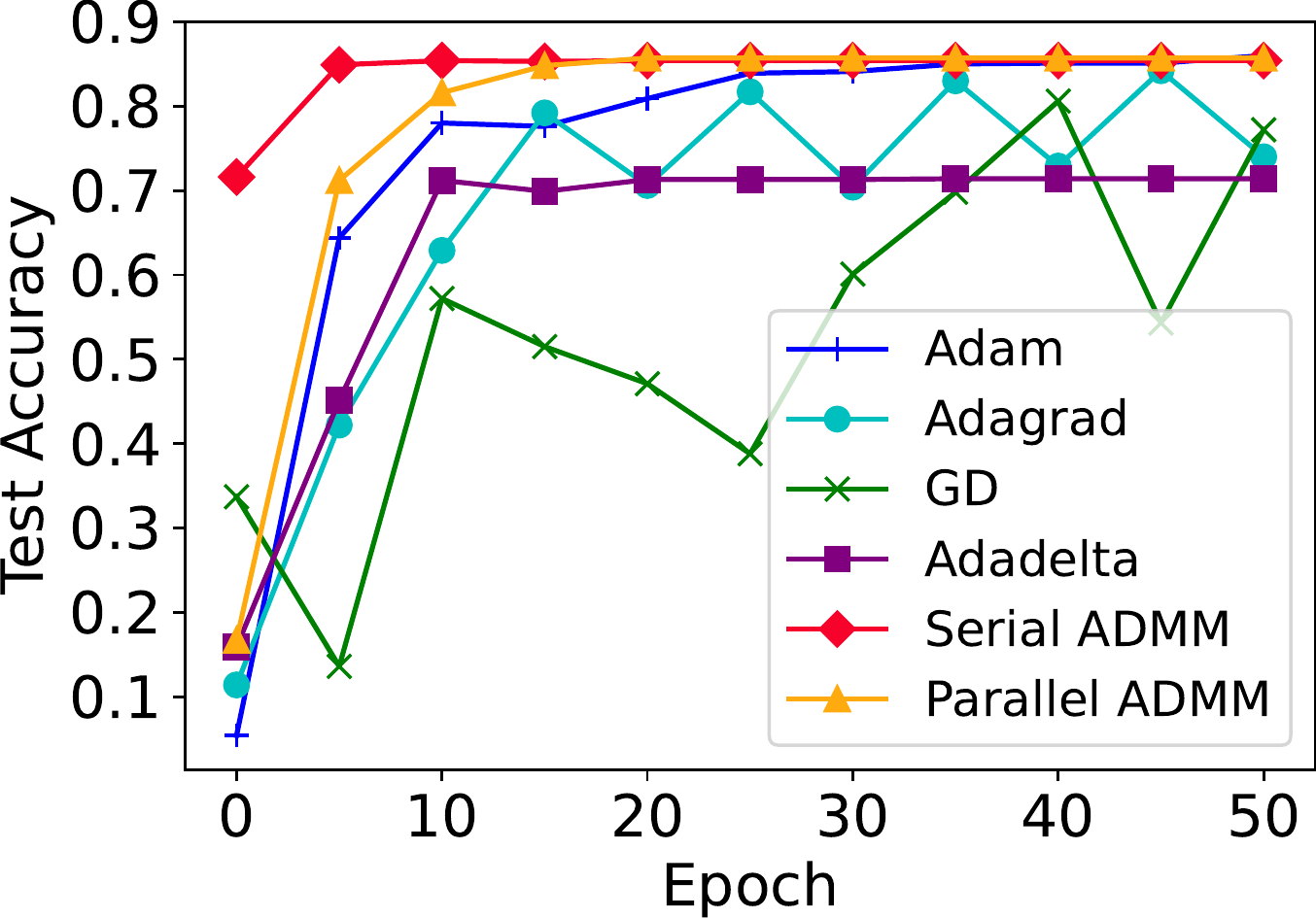}
        \centerline{\tabincell{c}{(c). Test accuracy\\ for Amazon Computers.} }
    \end{minipage}
    \hfill
    \begin{minipage}{0.24\linewidth}
    \includegraphics[width=\linewidth]{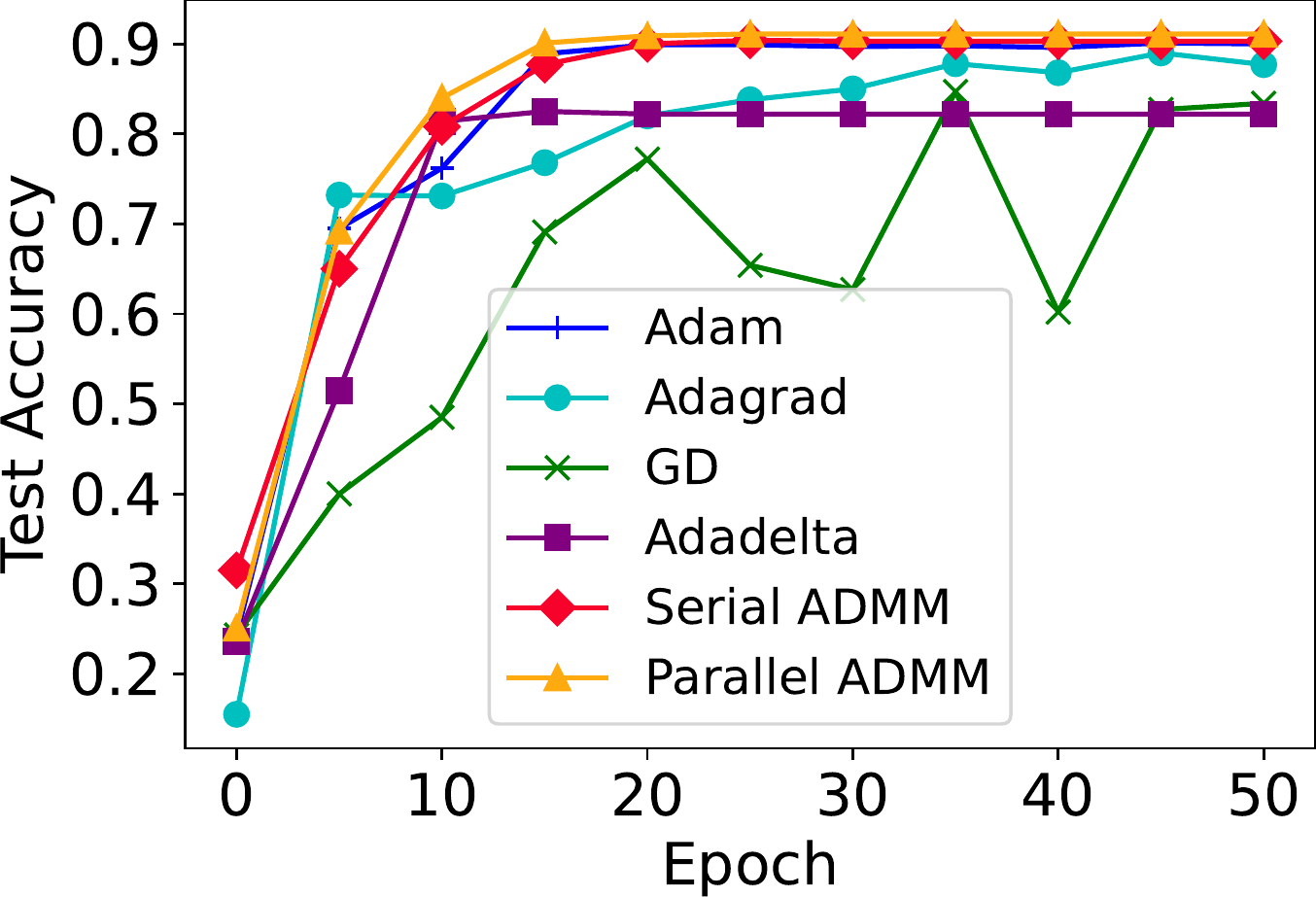}
        \centerline{\tabincell{c}{(d). Test accuracy\\ for Amazon Photo.} }
    \end{minipage}
    \caption{Training accuracy and test accuracy of all methods:  Serial ADMM algorithm and Parallel ADMM algorithm outperform most of  comparison methods in two datasets.}
    \label{fig:performance}
\end{figure*}

To validate the accuracy of the proposed community-based ADMM algorithms, we used the same GCN architecture and parameter settings for Serial ADMM and Parallel ADMM algorithms as those in Section \ref{sec:speedup}. SGD and its variants are state-of-the-art optimizers for GCN training and hence we used four of them as comparison methods, namely, Adaptive momentum estimation (Adam), Adaptive gradient algorithm (Adagrad), Gradient Descent (GD), and Adaptive learning rate method (Adadelta). For comparison methods, we used the following learning rate for Amazon Computers and Amazon Photo: $10^{-3}$ (Adam, Adagrad, and Adadelta) and $10^{-1}$ (GD) based on the optimal training performance.\\
\indent In this section, the accuracy of the proposed serial ADMM and parallel ADMM algorithms is analyzed against all comparison methods. Figure \ref{fig:performance} illustrates the training and test accuracy for all training methods on both datasets. The proposed Serial and Parallel ADMM algorithms reach the highest accuracy and outperform most comparison methods except for Adam, which perform almost the same compared to the proposed algorithms when epoch=$50$. Furthermore, the proposed two ADMM algorithms converge the fastest among all methods, and the convergence speed of Serial ADMM is ahead of that of Parallel ADMM in most situations.

\begin{table}[th!]
    \small
    \centering
    \begin{tabular}{c|c|c|c|c|c}
    \hline\hline
         Dataset&\tabincell{c}{ Node\#}&\tabincell{c}{Training  Sample\#}&\tabincell{c}{Test  Sample\#}&\tabincell{c}{ Class\#}&
{Feature\#}\\\hline 
\tabincell{c}{Amazon   Computers}&13752&1000&1000&10&767\\\hline 
\tabincell{c}{Amazon  Photo}&7650&800&1000&8&745\\\hline \hline
    \end{tabular}
    \vspace{-0.3cm}
    \caption{Two benchmark datasets.}
    \label{tab:dataset}
\end{table}
\vspace{-0.5cm}

\begin{table}[th!]
\small
\centering
\begin{tabular}{c|c|c|c|c|c}
\hline\hline
      &  Serial ADMM (sec) & \multicolumn{4}{c}{Parallel ADMM (sec)} \\\hline
   Dataset& Total & Training&Communication& Total&Speedup \\\hline
Amazon Computers & 80.82  &14.94 & 9.54     &24.48   & 3.30          \\\hline 
Amazon Photo  & 50.81     & 8.80     & 8.27 & 17.07 & 2.98       \\\hline \hline
\end{tabular}\caption{Comparison of training and communication time on two datasets.}\label{tab:time}
\end{table}
\section{Discussion and Future Work}
In this paper, we present the community-based ADMM algorithm to achieve both node parallelism and layer parallelism on training large-scale Graph Convolutional Networks (GCNs). Preliminary results on small benchmark datasets show that the community-based ADMM method leads to huge speedup and achieves excellent performance compared to state-of-the-art optimizers. However, its performance on large-scale datasets is still unsatisfactory. It is attributed to the relaxation of the problem (i.e. Problem \ref{prob: problem 2}). While problem relaxation realizes layerwise parallel training of GCN models, it may enlarge gaps between layers so that many solutions to the relaxed problem (i.e. Problem \ref{prob: problem 2}) do not fit the original problem (i.e. Problem \ref{prob: problem 1}). For small datasets, some of solutions to Problem \ref{prob: problem 2} still work on the Problem \ref{prob: problem 1}. But it is not the case for large-scale datasets. In the future, we may tackle this problem by exploring how to relax problems properly without loss of performance.
\bibliography{Reference}

\newpage
\appendix
\section{Update $Z_{l,m}^{k+1}$ }
\label{sec:update z}
The variable   $Z_{l,m}^{k+1}$ is updated as follows:
\begin{align*}
    Z_{l,m}^{k+1}\!\leftarrow\!\arg\!\min\nolimits_{Z_{l,m}}\! \mathcal{L}_\rho(\textbf{W}^{k+1},\textbf{Z},\textbf{U}^{k}).
\end{align*}

Notably, as output variables for intermediate layers  (i.e. $Z_{l}$) are involved in two constraints in Problem \ref{prob: problem 1}, updating $Z_{l,m}^{k+1}(l=1, \cdots, L-1)$ requires $\{\tilde{{A}}_{m,r}{Z}_{l,r}^k{W}_{l+1}^{k+1}\}$,  $\{\tilde{{A}}_{r,r'}{Z}_{l,r'}^k{W}_{l+1}^{k+1}\}$, $\{Z_{l+1,r}^k\}$, and $\{\tilde{{A}}_{m,r}Z_{l-1,r}^kW_{l}^{k+1}\}$, where $r \in \mathcal{N}_m\cup \{m\}$ and $r'\in \mathcal{N}_r\cup\mathcal{N}_m\backslash\{m\}$. In other words, the update of $Z_{l,m}^{k+1}$ requires information of second-order neighbors, which suffers from the bottleneck of the inherent neighbor explosion. 
To tackle this problem, the information from second-order neighbors can be conveyed via first-order neighbors, which are detailed as follows.

The first-order information is defined by:
\begin{equation*}
    p_{l,r\rightarrow m}^{k} \triangleq \tilde{{A}}_{m,r}Z_{l,r}^kW_{l+1}^{k+1} (l=0,\cdots,L-1).
\end{equation*}

The second-order information is defined in the following:
\begin{align}
\nonumber s_{l,r\rightarrow m}^{k} = [s_{l,r\rightarrow m}^{k, 1}, s_{l,r\rightarrow m}^{k, 2}]
    &\triangleq \begin{cases}
    [Z_{l+1,r}^k, \sum\nolimits_{r'\in \mathcal{N}_r\cup{\{r\}}\backslash\{m\}}\tilde{{A}}_{r,r'}Z_{l,r'}^kW_{l+1}^{k+1}] \\
    [Z_{L,r}^k - \sum\nolimits_{r'\in \mathcal{N}_r\cup{\{r\}}\backslash\{m\}}\tilde{{A}}_{r,r'}Z_{L-1,r'}^kW_{L}^{k+1}, U_r^k]
    \end{cases}\\
    &=\begin{cases}
    [Z_{l+1,r}^k, \sum\nolimits_{r'\in \mathcal{N}_r\cup{\{r\}}\backslash\{m\}}p_{l, r'\rightarrow r}^{k}] &l=0, \cdots,L-2\\
    [Z_{L,r}^k - \sum\nolimits_{r'\in \mathcal{N}_r\cup{\{r\}}\backslash\{m\}}p_{L-1, r'\rightarrow r}^{k}, U_r^k] & l=L-1.
    \end{cases}\label{eq:2nd order info}
\end{align}
We can see from \eqref{eq:2nd order info} that the second-order information forwarded by $r$ to $m$ can easily be constructed by community $r$ through aggregating its received first-order information $p_{l, r'\rightarrow r}^{k}$ from all $r' \in \mathcal{N}_r\cup{\{r\}}\backslash\{m\}$.
We further define 
$\mathbf{p}_{l, m}^{k} \triangleq \{p_{l,r\rightarrow m}^{k} | r \in \mathcal{N}_m\}$
and 
$ \mathbf{s}_{l, m}^{k} \triangleq \{s_{l,r\rightarrow m}^{k} | r \in \mathcal{N}_m\}.$
Then the objective for $Z_{l,m}^{k+1}$ can be modified as:
\small
\begin{align}
    \nonumber Z_{l,m}^{k+1}\!&\leftarrow\!\arg\!\min\nolimits_{Z_{l,m}}\frac{\nu}{2}\|{Z}_{l,m} - f_l(\sum_{r \in \mathcal{N}_m\cup{\{m\}}}p_{l-1,r\rightarrow m}^{k})\|_F^2\\
    \nonumber &+\frac{\nu}{2}\|{Z}_{l+1,m}^k - f_{l+1}(\tilde{{A}}_{m,m}{Z}_{l,m}{W}_{l+1}^{k+1}+\sum_{r \in \mathcal{N}_m}p_{l,r\rightarrow m}^{k})\|_F^2+\sum_{r \in \mathcal{N}_m }\!\frac{\nu}{2}\|s_{l,r\rightarrow m}^{k, 1}\!-\! f_{l+1}(\!\tilde{{A}}_{r,m}{Z}_{l,m}{W}_{l+1}^{k+1}+s_{l,r\rightarrow m}^{k, 2})  \|_F^2\\
    &\triangleq \psi({Z}_{l,m}, {Z}_{l+1,m}^k,{W}_{l+1}^{k+1}, \mathbf{p}_{l, m}^{k}, \mathbf{p}_{l-1, m}^{k},\mathbf{s}_{l, m}^{k})(l=1,\cdots,L-2),\label{eq:update z}
\end{align}
\normalsize
\small
\begin{align}
     \nonumber Z_{L-1,m}^{k+1} &\leftarrow \arg \min\nolimits_{Z_{L-1,m}} \frac{\nu}{2}\|{Z}_{L-1,m} -  f_{L-1}(\sum_{r \in \mathcal{N}_m\cup{\{m\}}}p_{L-2,r\rightarrow m}^{k})\|_F^2\\
     \nonumber&+\langle U_m^k, {Z}_{L,m}^k \!-\! ( \tilde{{A}}_{m,m}{Z}_{L-1,m}{W}_{L}^{k+1} \!+\! \sum_{r \in \mathcal{N}_m} p_{L-1,r\rightarrow m}^{k})\rangle
     \!+\! \frac{\rho}{2}\| {Z}_{L,m}^k \!-\!( \tilde{{A}}_{m,m}{Z}_{L-1,m}{W}_{L}^{k+1} \!+\! \sum_{r \in \mathcal{N}_m} p_{L-1,r\rightarrow m}^{k})\|_F^2\\
     \nonumber& + \sum_{r \in \mathcal{N}_m }  ( \langle s_{L-1,r\rightarrow m}^{k, 2},s_{L-1,r\rightarrow m}^{k, 1}-   \tilde{{A}}_{r,m}{Z}_{L-1,m}{W}_{L}^{k+1}\rangle  + \frac{\rho}{2}\|s_{L-1,r\rightarrow m}^{k, 1}-   \tilde{{A}}_{r,m}{Z}_{L-1,m}{W}_{L}^{k+1} \|_F^2)\\
     & \triangleq \psi({Z}_{L-1,m},  {Z}_{L,m}^k, {W}_{L}^{k+1}, \mathbf{p}_{L-1, m}^{k}, \mathbf{p}_{L-2, m}^{k},\mathbf{s}_{L-1, m}^{k}),\label{eq:update zl_1}
\end{align}
\normalsize
and 
\small
\begin{align}
    \nonumber Z_{L,m}^{k+1} &\leftarrow \arg \min\nolimits_{Z_{L,m}}  R(Z_{L,m},Y_m)
     \!+\!\langle U_m^k,{Z}_{L,m} - (\tilde{{A}}_{m,m}{Z}_{L-1,m}^k{W}_{L}^{k+1}+\sum_{r \in \mathcal{N}_m}p_{L-1,r\rightarrow m}^{k}) \rangle\\
     \nonumber &+\frac{\rho}{2}\|{Z}_{L,m} - (\tilde{{A}}_{m,m}{Z}_{L-1,m}^k{W}_{L}^{k+1}+\sum_{r \in \mathcal{N}_m}p_{L-1,r\rightarrow m}^{k})\|_F^2\\
     & \triangleq \psi( Z_{L,m}, Z_{L-1,m}^k, {W}_{L}^{k+1}, \mathbf{p}_{L-1, m}^{k}, U_m^k).\label{eq:update zl}
\end{align} 
\normalsize

Obviously, $Z_{l,m}^{k+1}$ for different $m$ and $l$ can all be updated in parallel. Furthermore, community $m$ should receive $p_{l,r\rightarrow m}^{k} (l<L)$ and $s_{l,r\rightarrow m}^{k}(l<L)$ from all its neighbor communities $r$ before updating $Z_{l,m}^{k+1}$.
In addition, the close-form solution to ${Z}_{l,m}^{k+1} (l<L)$  requires time-consuming matrix inverse operation. Similar to update $W$, the quadratic approximation technique is applied as follows:
\small
\begin{equation}\label{z_qa_1}
     {Z}_{l,m}^{k+1}\leftarrow \arg\min\nolimits_{{Z}_{l,m}} Q_l({Z}_{l,m}; \theta_{l,m}^{k+1}),
\end{equation}
\normalsize

where
\small
\begin{align*}
Q_{l,m}({Z}_{l,m}; \theta_{l,m}^{k+1})&\triangleq\psi({Z}_{l,m}^k,  {Z}_{l+1,m}^k, {W}_{l+1}^{k+1}, \mathbf{p}_{l, m}^{k}, \mathbf{p}_{l-1, m}^{k},\mathbf{s}_{l, m}^{k})+\frac{\theta_{l,m}^{k+1}}{2}\|{Z}_{l,m}-{Z}_{l,m}^k\|_F^2\\
&\!+\! \langle\nabla_{{Z}_{l}^k} \psi({Z}_{l,m}^k,  {Z}_{l+1,m}^k,{W}_{l+1}^{k+1}, \mathbf{p}_{l, m}^{k}, \mathbf{p}_{l-1, m}^{k},\mathbf{s}_{l, m}^{k}),{Z}_{l,m}-{Z}_{l,m}^k\rangle,l<L,\\
\end{align*}
\normalsize
and $\theta_{l,m}^{k+1}>0$ is a parameter that should satisfy:
\small
\begin{align}\label{z_cri}
 Q_{l,m}({Z}_{l,m}^{k+1}; \theta_{l,m}^{k+1}) \geq
  & \psi({Z}_{l,m}^{k+1},  {Z}_{l+1,m}^k, {W}_{l+1}^{k+1}, \mathbf{p}_{l, m}^{k}, \mathbf{p}_{l-1, m}^{k},\mathbf{s}_{l, m}^{k}),l<L
\end{align}
\normalsize
The solution is:
\small
\begin{align}
{Z}_{l,m}^{k+1}
\!\leftarrow{Z}_{l,m}^{k}\!-\!\nabla_{{Z}_{l,m}^k}\!\psi({Z}_{l,m}^k,  {Z}_{l+1,m}^k, {W}_{l+1}^{k+1}, \mathbf{p}_{l, m}^{k}, \mathbf{p}_{l-1, m}^{k},\mathbf{s}_{l, m}^{k})/\theta_{l,m}^{k+1},l<L.
\end{align}
\normalsize
Finally, \eqref{eq:update zl} (i.e. $l=L$) can be solved directly via Fast Iterative Soft-Thresholding Algorithm (FISTA) \cite{Beck2009}.
\end{document}